\documentclass[letterpaper, 10 pt, conference]{ieeeconf}  % Comment this line out if you need a4paper

\IEEEoverridecommandlockouts                              % This command is only needed if 
\usepackage{graphics} % for pdf, bitmapped graphics files
\usepackage{epsfig} % for postscript graphics files
\usepackage{amsmath} % assumes amsmath package installed
\usepackage{amssymb}  % assumes amsmath package installed

\usepackage{graphicx}
\usepackage{steinmetz}
\usepackage{caption}
\usepackage{subcaption}

\title{\LARGE \bf
Learning Moving-Object Tracking with FMCW LiDAR
}

\author{Yi Gu$^{1}$, Hongzhi Cheng$^{1}$, Kafeng Wang$^{2}$, Dejing Dou$^{3}$, Chengzhong Xu$^{1 *}$ and Hui Kong$^{1 *}$% <-this % stops a space
\thanks{The Doppler LiDAR data were provided by Guangshao Technology.}% <-this % stops a space
\thanks{$^{*}$Corresponding author.}%
\thanks{$^{1}$Yi Gu, Hongzhi Cheng, Hui Kong and Chengzhong Xu are with Faculty of Science and Technology, 
        University of Macau, Taipa, Macau, China.
        {\tt\small \{yc17463; hongzhicheng; huikong; czxu\}@um.edu.mo}}%
\thanks{$^{2}$Kafeng Wang is with Shenzhen Institute of Advanced Technology, Chinese Academy of Sciences and University of Chinese Academy of Sciences.
        {\tt\small kf.wang@siat.ac.cn}}%
\thanks{$^{3}$Dejing Dou is with Baidu Research.}
}
\UseRawInputEncoding

\begin{document}

\maketitle
\thispagestyle{empty}
\pagestyle{empty}

%%%%%%%%%%%%%%%%%%%%%%%%%%%%%%%%%%%%%%%%%%%%%%%%%%%%%%%%%%%%%%%%%%%%%%%%%%%%%%%%
\begin{abstract}
% Using LiDAR to tackle Multi-Object Tracking and Segmentation (MOTS) problem is less explored compared to that in the image domain due to its difficulty in processing high-dimension data across time. 
% Due to the fact that only spatial information is available, most of the existing commercial LiDAR sensors have problems in tracking moving objects (e.g., vehicles). In point clouds, moving objects usually do not exhibit uniform patterns, and tracking performance would decline further with static points confused in. Although some end-to-end deep learning algorithms can achieve remarkable improvements over traditional handcraft feature learning methods, it is inevitable to label a large amount of data in most of these methods.

In this paper, we propose a learning-based moving-object tracking method utilizing our newly developed LiDAR sensor, Frequency Modulated Continuous Wave (FMCW) LiDAR. Compared with most existing commercial LiDAR sensors, our FMCW LiDAR can provide additional Doppler velocity information to each 3D point of the point clouds. Benefiting from this, we can generate instance labels as ground truth in a semi-automatic manner. Given the labels, we propose a contrastive learning framework, which pulls together the features from the same instance in embedding space and pushes apart the features from different instances, to improve the tracking quality. Extensive experiments are conducted on our recorded driving data, and the results show that our method outperforms the baseline methods by a large margin.

%In this paper, we propose a learning-based moving-object tracking method using our newly developed LiDAR sensor, Frequency Modulated Continuous Wave (FMCW) LiDAR. Compared with most existing commercial LiDAR sensors, our FMCW LiDAR can provide additional Doppler velocity information to each 3D point of the point clouds. Benefiting from this, we can annotate our data in an effortless semi-automatic manner to provide the ground-truth instance-level labels. Given the labels, we propose a supervised contrastive learning to pull together the features from the same instance in embedding space and push apart the features from different instances. Extensive experiments are conducted on our recorded driving data, and the results show that our method outperforms the baseline methods by a large margin.

\end{abstract}

%%%%%%%%%%%%%%%%%%%%%%%%%%%%%%%%%%%%%%%%%%%%%%%%%%%%%%%%%%%%%%%%%%%%%%%%%%%%%%%%
\section{INTRODUCTION}
Multi-Object Tracking (MOT) is a significant module in modern autonomous driving pipelines. As one of the key tasks of MOT, moving-object tracking can strongly affect the local planning strategy of the ego-vehicle. With this consideration, we only focus on tracking moving objects in this work. In the literature, most state-of-the-art (SOTA) MOT algorithms are based on deep learning techniques, and require detection or segmentation results. For example, the recent MOT methods \cite{yin2021center, zeng2021cross} require to detect objects first, and then associate detections across frames. The training processes of these methods need the bounding box labels for the moving targets in each LiDAR frame. The 4D Panoptic LiDAR Segmentation (4D-PLS) \cite{aygun20214d} is based on a tracking-by-segmentation paradigm and has shown remarkable performance in the LiDAR-based MOT task. However, the labeling of the segmentation task is more expensive, even with many advanced tools \cite{bloembergen2021automatic}.

Basically, these methods require a large amount of manual annotations to train a deep neural network in a supervised way, and may not generalize well in scenes beyond the training set. Typically, there are two strategies to overcome this problem. One strategy is to investigate how to utilize data efficiently or how to train a model with less supervision, such as self-supervised learning \cite{he2020momentum, he2021masked}, weakly supervised learning \cite{zhou2018brief}, few-shot learning \cite{snell2017prototypical}, etc.

The other potential strategy to solve this problem is from sensor innovation or fusion perspective, since it can usually provide extra information acting as a kind of self-supervision during model training. Our work belongs to this route, where we utilize our newly developed FMCW LiDAR to track moving objects in a deep learning paradigm. Our target is to accomplish this task with as little manual-labeling effort as possible.

\begin{figure}
     \centering
        %  \hfill
     \begin{subfigure}[b]{0.235\textwidth}
         \centering
         \includegraphics[width=\textwidth]{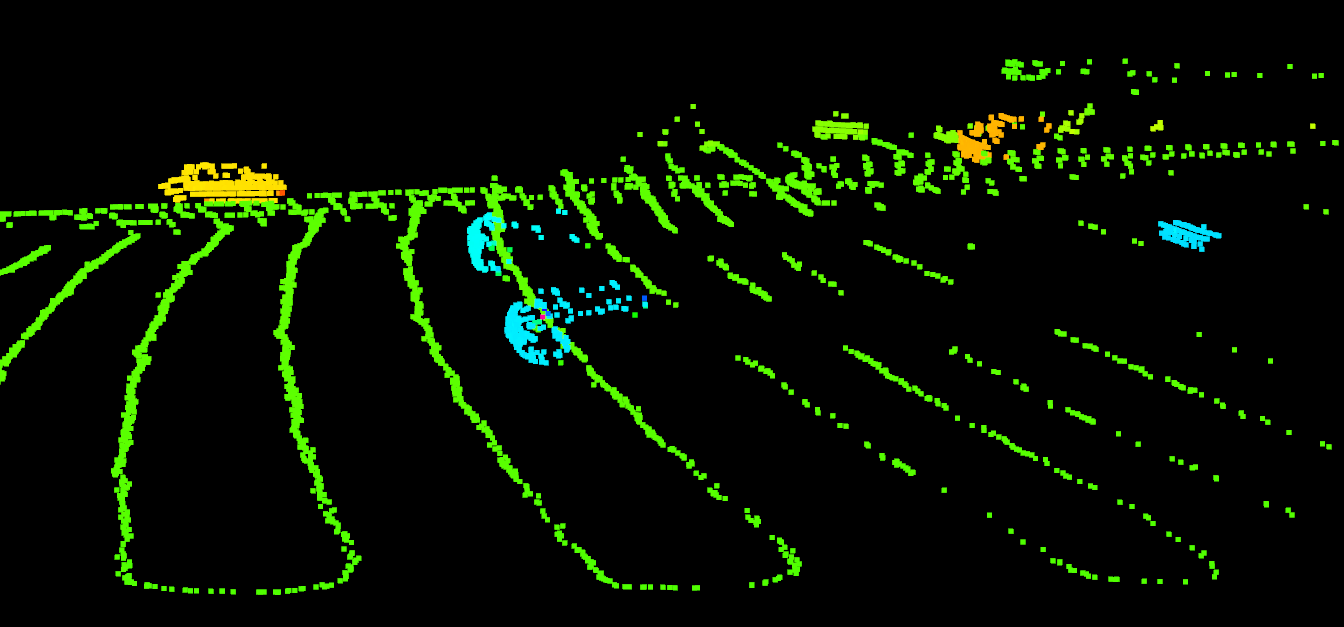}
         \captionsetup{font={footnotesize}} 
         \caption{Point clouds with velocity}
         \label{fig:b}
     \end{subfigure}
      \begin{subfigure}[b]{0.235\textwidth}
         \centering
         \includegraphics[width=\textwidth]{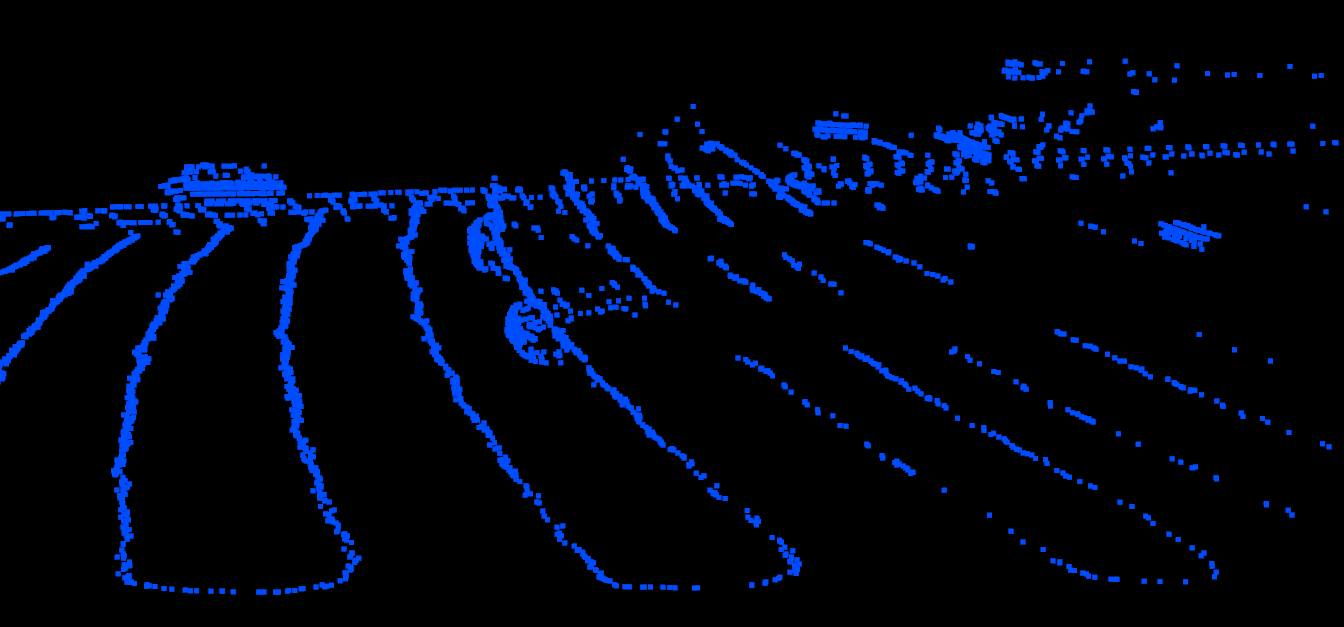}
         \captionsetup{font={footnotesize}} 
         \caption{Original 3D point clouds.}
         \label{fig:a}
     \end{subfigure}
     \captionsetup{font={footnotesize}} 
        \caption{Visual comparison of the FMCW LiDAR measurements and the range-only measurements. We use the HSI color model to visualize the velocity. Hue values in (a) are arranged by the Doppler velocity. A lower hue indicates faster speed towards the sensor.}
        \label{fig:2 graphs}
\end{figure}

The major difference of FMCW LiDAR from most existing LiDAR sensors lies in its supplement of speed information in addition to the 3D coordinates of the point cloud. Its basic principle is based on the Doppler effect that can provide the measurement of relative velocity between each sensed 3D world point to the FMCW LiDAR sensor along the radial direction. Figure \ref{fig:2 graphs} shows a certain frame of FMCW LiDAR point cloud with and without velocity information, respectively.

In contrast to the above-mentioned methods, the advantages of using the FMCW LiDAR are reflected in two aspects. The first one is to reduce the cost of data labeling, and the other is to provide the motion cue for data association.

\begin{figure}
     \centering
        %  \hfill
     \begin{subfigure}[b]{0.235\textwidth}
         \centering
         \includegraphics[width=\textwidth]{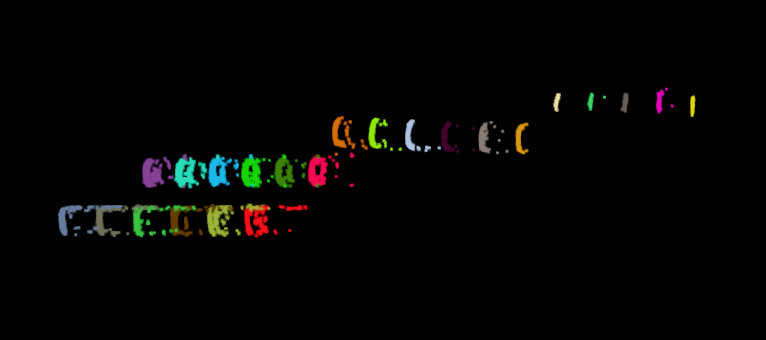}
         \captionsetup{font={footnotesize}} 
         \caption{}
         \label{da}
     \end{subfigure}
      \begin{subfigure}[b]{0.235\textwidth}
         \centering
         \includegraphics[width=\textwidth]{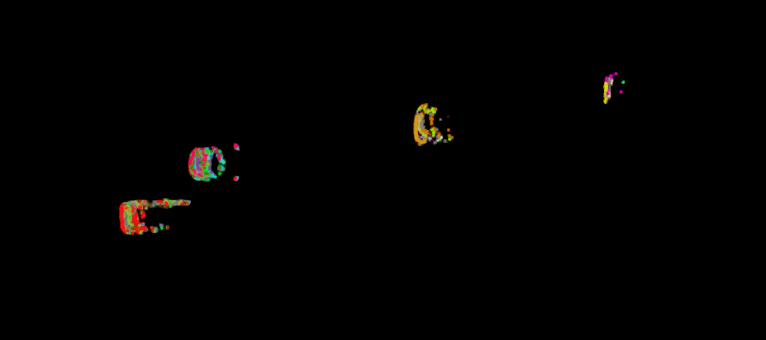}
         \captionsetup{font={footnotesize}} 
         \caption{}
         \label{db}
     \end{subfigure}
     \captionsetup{font={footnotesize}} 
        \caption{The illustration of the position compensation. (a) is the result of clustering in each frame and aligned in the current frame. (b) is the result of the position compensation according to the velocity of each cluster.}
        \label{data-association}
\end{figure}

In our work, we are devoted to fully utilizing these two advantages of the FMCW LiDAR. The first stage of our method is data labeling, a semi-automatic annotation process. Specifically, we initially separate static points from dynamic ones with velocity information. Within each LiDAR frame, a clustering algorithm \cite{ester1996density} is applied to the moving points to get each individual moving cluster. Then data association is carried out on the moving clusters with estimated ego-motion and position (velocity) compensation. 
As shown in Figure \ref{da}, with the estimated ego-motion information (by a LiDAR SLAM algorithm for our case), the segmented moving clusters across multiple frames can be aligned to a reference frame (set as the current frame). As shown in Figure \ref{db}, based on position compensation with velocity information on each cluster, we can further align each cluster so that the clusters corresponding to the same object are aggregated, where the initial data association can be achieved based on the distance between the centers of the aggregated clusters. %If the distance is within the fixed threshold and the two clusters are the nearest pair, we treat them as tracked. 
It is observed that data association can be established for most cases in this way. But the density parameter of clustering and the distance threshold need to be tuned scene-by-scene. We annotate our dataset by manually tuning these hyper-parameters. After that, each point is assigned an instance ID without semantic class label. Note that the cost of this semi-automatic annotation process is far less than the per-point annotations and bounding box annotations.

With the above ground-truth labels, we can train an end-to-end model to track the moving objects. Generally, it is not robust to track objects with only instance-level supervision. We propose a contrastive learning framework to make the features belonging to the same instance close and the features from different instances apart. With these high-quality features, we can track the instances across time simply by calculating the association probabilities of their points.

In summary, our main contributions are of two folds:
\begin{itemize}
	\item We present the first deep learning based method on moving-object tracking with the FMCW LiDAR. We propose a contrastive learning framework to make full use of the instance-only supervision.
	
	\item Our semi-automatic annotation method with FMCW LiDAR can significantly reduce the labeling costs, which can motivate more related works based on newly developed sensors.
\end{itemize}

\section{RELATED WORK}
\begin{figure*}[ht]
\centering
{\includegraphics[width=0.96\textwidth]{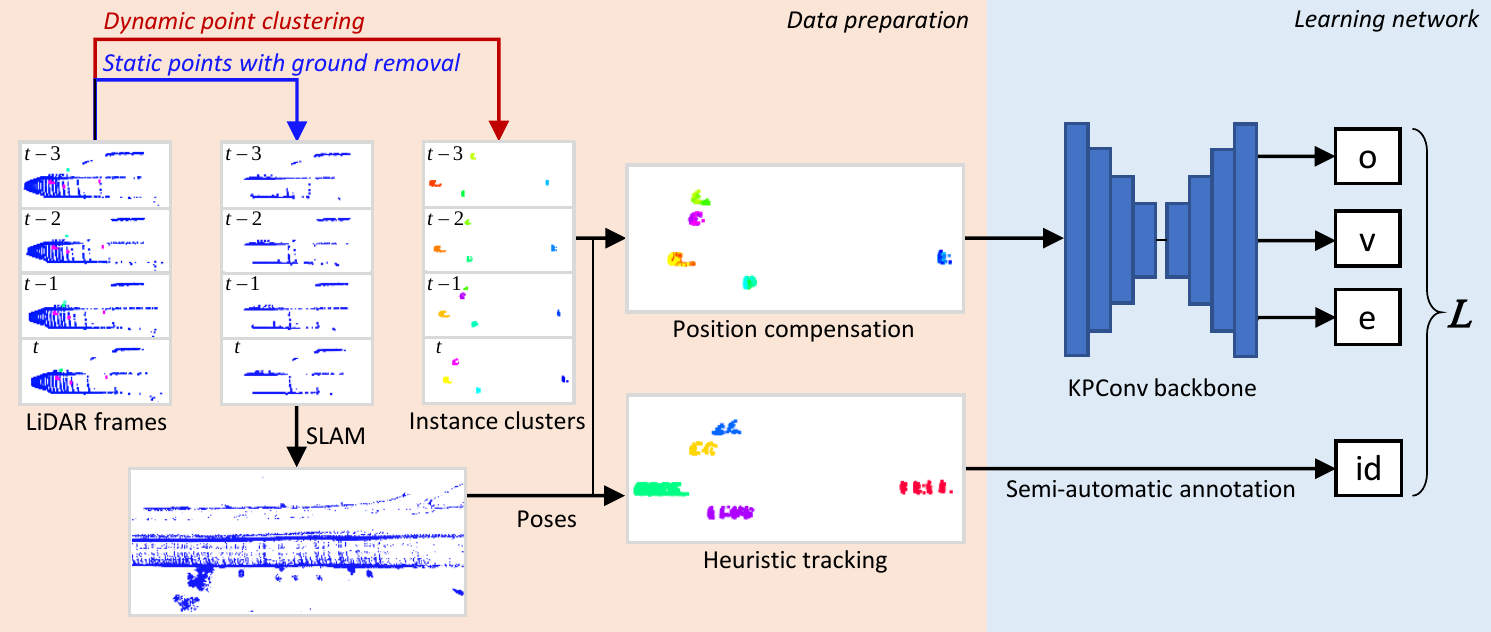}}
\captionsetup{font={footnotesize}} 
\caption{An overview of our moving-object tracking pipeline (window size $\tau = 4$ for illustration). } 
\label{net}   
\end{figure*}
\subsection{FMCW LiDAR}
FMCW LiDAR is a newly developed type of ranging sensors and was first researched in wind measurements \cite{Dopplerfirst}. Compared with other popular LiDAR schemes, the scheme of FMCW LiDAR makes it able to measure the relative velocity (Doppler velocity) of each scanned point with respect to the sensor along the radial direction by the Doppler effect \cite{hexsel2022dicp}. Since this kind of LiDAR sensors is seldom available, there are few related works specifically on perception. Recently, DICP \cite{hexsel2022dicp} presented a novel method for point cloud registration by using Aeva’s Aeries I FMCW LiDAR. Guo et al. \cite{guo2021doppler} proposed a clustering and velocity estimation method based on Doppler velocity with CARLA \cite{dosovitskiy2017carla} simulation. Jie Shan et al. \cite{ma2019moving, peng2021detection} proposed detection and tracking methods based on hand-crafted features with Kalman filter to estimate the state of moving objects by using Blackmore FMCW LiDAR. Their methods are not based on deep learning, simple and efficient, but in low accuracy.

\subsection{3D Multi-Object Tracking}
The dominant methods in this field exploit the tracking-by-detection paradigm, which detects objects first, then associates detections across time. The association step is usually formulated as a bipartite matching problem, where the key to matching relies on the appearance and motion models. Most of the LiDAR-based tracking methods hinge on the motion for association due to the lack of the texture cue. For example, CenterPoint \cite{yin2021center} achieved this by estimating the velocity of each object. Other methods \cite{chiu2021probabilistic, zeng2021cross} exploited appearance information from camera to enhance the association quality. SimTrack \cite{luo2021exploring} avoided heuristic matching step by joint detection and tracking in an end-to-end manner. It detects objects in bird-eye-view and predicts the offset of each object directly. Its association step can be finished with a simple lookup operation. However, all these methods require bounding box annotations for detection.

MOTS \cite{voigtlaender2019mots} proposed the multi-object tracking and segmentation task for the first time and MOTP \cite{hurtado2020mopt} extended it to panoptic segmentation and 3D domain. 4D-PLS \cite{aygun20214d} explored tracking-by-segmentation paradigm, which can associate objects at point level. The core of 4D-PLS is the density-based clustering, which benefits from their semantic features. Our method can be sorted into this category. But in our dataset, there are only instance annotations without semantic labeling, which is more challenging than previous works.

\subsection{Contrastive Learning}
Contrastive learning based methods have dominated self-pretraining in vision in recent years \cite{he2020momentum, chen2020simple}. The main idea of contrastive learning is the following: pull together an anchor and a "positive" sample in embedding space, and push apart the anchor from many "negative" samples \cite{khosla2020supervised}. In the self-supervised learning domain, there are no labels available. The positive samples are usually drawn from data augmentations, and the negative pairs are formed by the anchor and remaining samples in the mini-batch. When labels are available, a similar idea can also be applied in the supervised learning scheme with different loss functions, such as Triplet Loss \cite{weinberger2009distance, schroff2015facenet}, N-pairs Loss \cite{sohn2016improved} and Supervised Contrastive Loss \cite{khosla2020supervised}. Recently, Marcuzzi et al. \cite{marcuzzi2022contrastive} proposed to use contrastive learning to make instance association, which is similar to our work. The difference is that we only utilize the instance information, while their method also requires semantic information.

\section{Method}
The goal of our approach is to track the moving objects across different times. To achieve this, we first annotate our data by Heuristic Tracking (See Sec. \ref{A}), which is a semi-automatic process and can provide us high-quality labeled data as ground truth. Thereafter, we train a deep neural network to achieve better performance (See Sec. \ref{B}). Due to lack of the supervision from semantic class labels, we utilize the contrastive learning method to improve the effectiveness of the association (See Sec. \ref{C}). Figure \ref{net} is the overview of our pipeline.

\subsection{Heuristic Tracking}\label{A}
% \subsection{Semi-Automatic Annotation}\label{A}
We first filter out the abnormal points such as geometry outliers and those with unrealistic velocity values. Then we separate dynamic and static points apart according to the ego velocity $V_{car}$. As this value is not recorded in our dataset, we estimate it by the mean velocity of the ground points in front view, noted as $V_g$, which is theoretically equal to $-V_{car}$. The ground points are extracted via the RANSAC \cite{fischler1981random} algorithm. We treat the points whose velocities are within $V_g\pm V_m$ as static points, where $V_m$ is a constant value and is set to 0.2 $m/s$ in our experiments. Then the remaining points are dynamic points. Henceforth, we only focus on these moving points.

Since there are no bounding boxes nor instance segmentation masks, we need to define the concept of the object first. We apply DBSCAN \cite{ester1996density} to cluster the moving points so that each cluster can be treated as an object. Though most of our cases can be successfully clustered, there are still some objects that failed to be separated out, which need to be further tuned by adjusting the density parameter of DBSCAN.

To track across the frames, we compensate the position of each cluster with the mean velocity of its own points. More specifically, for a window of $\tau$ frames at times $\{max(0, t+1-\tau), ..., t\}$, we first align them to the current frame $t$ according to the ego-motion estimation results provided by a SLAM approach \cite{zhang2014loam}. As we operate in consecutive frames and the window time duration in our experiments is less than 1 $s$, we can treat the motion of each object as uniform motion during this period. In this way, we can compensate the position of each cluster by computing the relative displacement and translating the corresponding points to simplify the association, as shown in Figure \ref{data-association}. Ideally, the points from the same object instance can be aggregated together so that we can do association by nearest neighbor query, accompanied by a distance threshold, $d_N$. That is to say, if two clusters at different times are nearest neighbors after shifting and the distance between them is within the threshold $d_N$, we consider them as the same instance. To make this operation more robust, we represent the position of each cluster by its centroid in the bird-eye-view to eliminate the height error.

\subsection{4D-PLS Review}\label{B}
We annotate our data based on Heuristic Tracking (See \ref{annotation} for details). With the obtained ground-truth labels, we are able to train a neural network to further improve the tracking quality. We build up our model based on the 4D-PLS \cite{aygun20214d} architecture. We will review the 4D-PLS first in this subsection and introduce our modifications in the next one.

The input of the 4D-PLS is a 4D volume, which contains consecutive frames of point clouds in a time window, aligned to the current frame. The main component of the model is the density-based clustering by the following equation,
\begin{equation}
    \hat{p}_{ij}=\frac{1}{(2\pi)^{\frac{D}{2}}|\Sigma_i|^{\frac{1}{2}}}exp(-\frac{1}{2}(e_i-e_j)^{T}\Sigma_i^{-1}(e_i-e_j)),
\end{equation}
where $\hat{p}_{ij}$ is the probability of the two points $p_i$ and $p_j$ belonging to the same instance. $e_i\in\mathbb{R}^{D}$ is the embedding of the $p_i$. Note that in this formulation, $p_i$ should be the instance center approximation. So $\Sigma_i$ is the variance matrix of $p_i$. 
% $D$ is the dimension of the embedding $e$.

Therefore, the outputs of the 4D-PLS \cite{aygun20214d} are semantic class, objectness, variance and point embedding. Here the objectness is a scalar used to assess how close each point is to its instance center. 

\subsection{Contrastive Moving-Object Tracking}\label{C}
There are two drawbacks if we directly use the original 4D-PLS \cite{aygun20214d}. The first one lies in the window size limitation. Longer temporal windows can aggregate more information, while the Gaussian assumption is only valid for shorter temporal windows in 4D-PLS. Besides, since we discard the static points and only focus on the moving points, a small window size is not efficient. Thanks to the superiority of velocity information, this problem can be solved simply by the position compensation described before. After that, the Gaussian assumption can be valid for longer temporal windows. We validate this conclusion in our ablation studies.

Another problem is that our dataset has no bounding box labels nor semantic labels needed in 4D-PLS, and only the above obtained instance IDs can be used for supervised learning. It is hard to associate the points at different times without a strong semantic cue. We tried to apply 4D-PLS with only instance level supervision, where the objectness loss and instance loss are defined as:
\begin{equation}
    L_{obj} = \sum\limits_{i=1}^{N}({\hat{o}_{i}}-{o}_{i})^2, \quad \hat{o}_i, o_i \in [0, 1]
\end{equation}
\begin{equation}
\begin{aligned}
L_{ins} =  \sum\limits_{j=1}^{K}\sum\limits_{i=1}^{N}({\hat{p}_{i j}}-p_{i j})^2, \quad p_{i j}=\begin{cases}1,&\mbox{if $p_{i} \in {I_j}$};\\0,&\mbox{otherwise}.\end{cases}
\end{aligned}
\end{equation}
In the above losses, notations are defined as follows: N is the number of points; $K$ is the number of instances in the current time window; $I_j$ represents the point set of the $j$-th instance; ${\hat{o}_{i}}$ is the approximation to the instance objectness ${o}_{i}$ after position compensation. Note that since the clusters from the same instance have been aggregated, we obtain the ground-truth objectness according to the instance center, while in 4D-PLS, it is calculated for each cluster.

This method does not work well in our experiments (See Table \ref{abl_table_in} and Figure \ref{results} for better comparison). Due to the lack of ground-truth class labels, we cannot rely on semantic information to obtain the class related features, which are considerably helpful for the association step.

Besides the motion cue, the appearance of the instance is also an important cue that can be explored further \cite{luo2021exploring, zeng2021cross}. It means that the clusters from the same instance should be similar, while from different instances should be distinct. Inspired by this hint, we propose to train a class-agnostic model in a contrastive learning paradigm \cite{he2020momentum, khosla2020supervised}. More specifically, we make use of the supervised contrastive learning \cite{khosla2020supervised} loss to further utilize the instance ID information. We represent each instance at each timestamp, i.e., each cluster, by its global feature $z_i$. Following Pointnet \cite{qi2017pointnet}, we obtain this global feature $z_i$ by applying a symmetric function on all embeddings of its points:
\begin{equation}
    z_i = f(x_{i1}, ..., x_{in}) \approx g(e_{i1}, ..., e_{in}),
\end{equation}
where $x_{ij}$ is the $j$-th point of the cluster $i$ and $e_{ij}$ is its corresponding feature embedding. We define that the positive samples are clusters from the same instance and the negative samples are those from different instances.

After that, we can use the following expression to compute the loss:
\begin{equation}
    L_{SC} = \sum\limits_{i\in N}\frac{-1}{|P(i)|}\sum\limits_{p\in P(i)}log\frac{exp(z_i \cdot z_p/\tau)}{\sum\limits_{a\in A(i)}exp(z_i \cdot z_a/\tau)},
\end{equation}
where $N$ is the number of clusters. $A(i)$ is the index set of all the clusters within the window except $i$. $P(i)$ is the index set of the clusters from the same instance as $i$.

As for inference, we follow the 4D-PLS to associate points in two stages. For the first stage, we select the point $p_i$ which has the highest objectness score as the instance center proxy. Then, we do the density-based clustering for all candidate points. If $p_{ij} > p_{threshold}$, we assign it to the instance of $p_i$ and remove it from candidate points. These steps are repeated until all points are clustered. For the second stage, we perform cross-volume association greedily based on the overlap score. When the overlap is below a threshold, we assign a new ID.

\section{Experiments}
\subsection{Dataset}
We utilize the latest Guangshao 2.0 FMCW LiDAR to collect our data. The horizontal field-of-view of our LiDAR is $37.5^{\circ}$, and the vertical field-of-view is $16.7^{\circ}$. The maximum perceived range is $300$ $m$. The precision of the velocity measurement is $5$ $cm/s$. The sampling rate is $10$ $Hz$. The data were collected in seven different scenes, 5 from highway and 2 from urban scenes. Each scene contains 300-500 frames after sampling and all data contain 3200 frames in total. We use 2400 scannings as the training set and the left as the test set.
\begin{figure*}[htbp]
     \centering
    %  \begin{subfigure}[b]{0.225\textwidth}
    %      \centering
    %      \includegraphics[width=\textwidth]{res/res_flip.png}
    %      \caption{}
    %  \end{subfigure}
    %       \begin{subfigure}[b]{0.225\textwidth}
    %      \centering
    %      \includegraphics[width=\textwidth]{res/res_flip.png}
    %      \caption{}
    %  \end{subfigure}
    %       \begin{subfigure}[b]{0.225\textwidth}
    %      \centering
    %      \includegraphics[width=\textwidth]{res/res_flip.png}
    %      \caption{}
    %  \end{subfigure}
    %       \begin{subfigure}[b]{0.225\textwidth}
    %      \centering
    %      \includegraphics[width=\textwidth]{res/res_flip.png}
    %      \caption{}
    %  \end{subfigure}
     \begin{subfigure}[b]{0.24\textwidth}
         \centering
         \includegraphics[width=\textwidth]{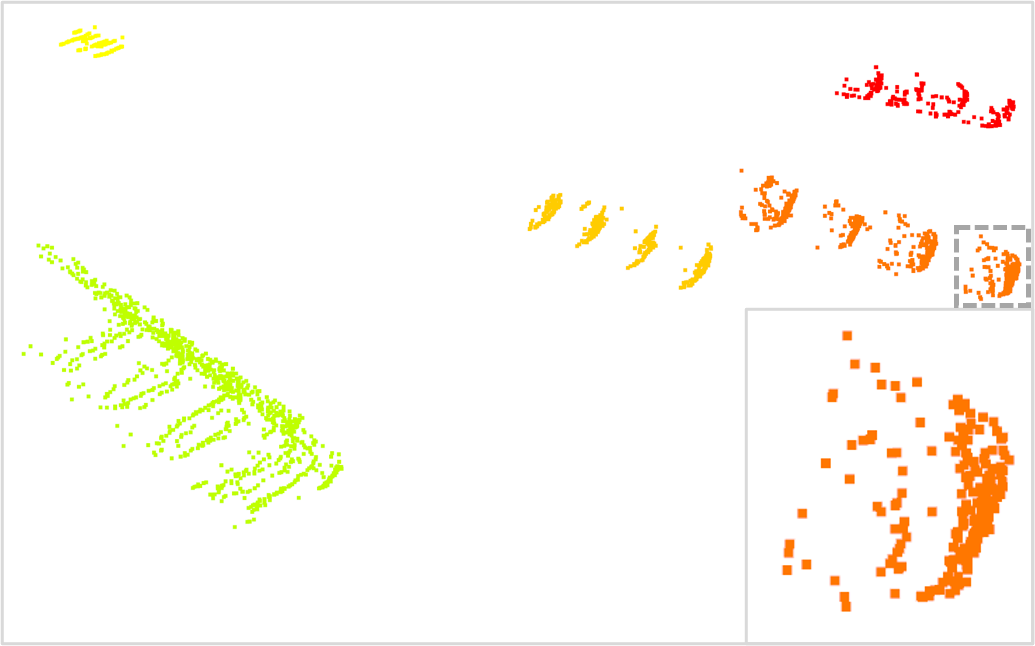}
         \hfill
          \includegraphics[width=\textwidth]{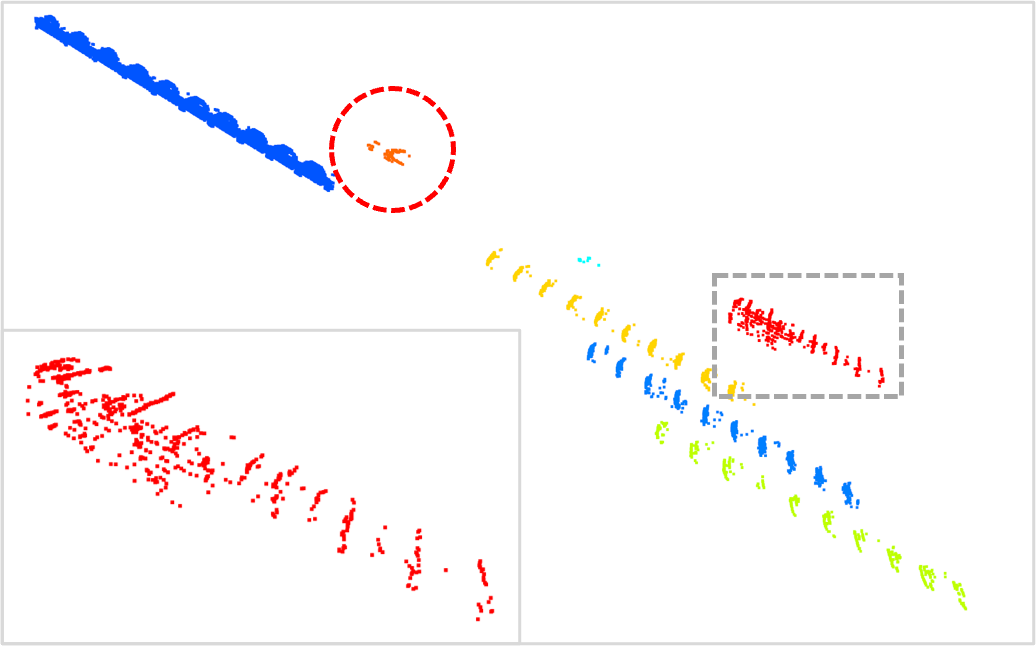}
          \captionsetup{font={footnotesize}} 
         \caption{Ground Truth}
     \end{subfigure}
     \begin{subfigure}[b]{0.24\textwidth}
         \centering
         \includegraphics[width=\textwidth]{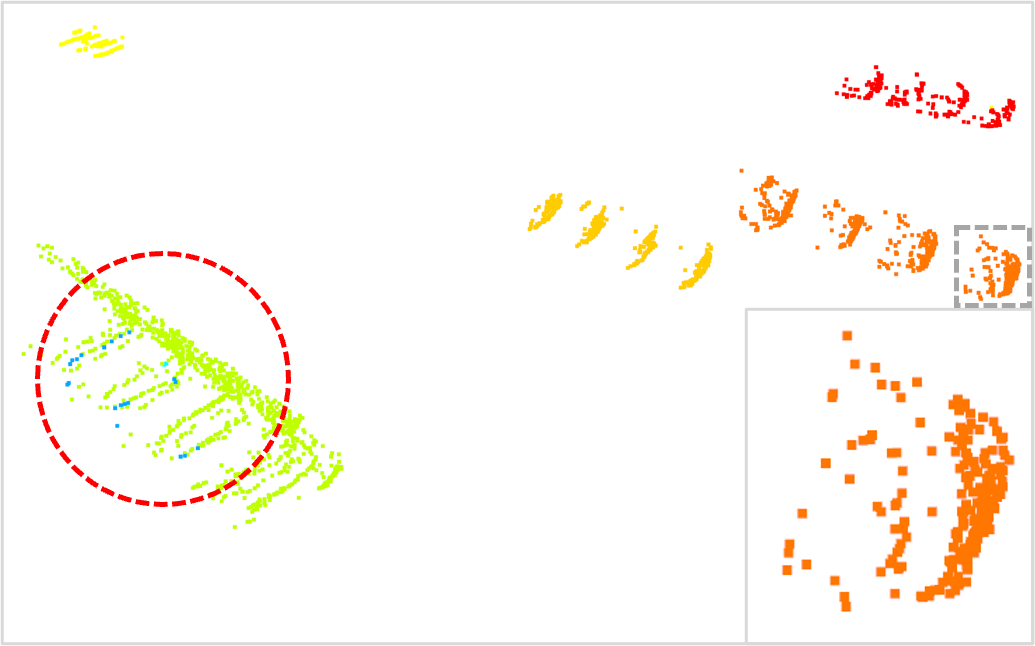}
         \hfill
         \includegraphics[width=\textwidth]{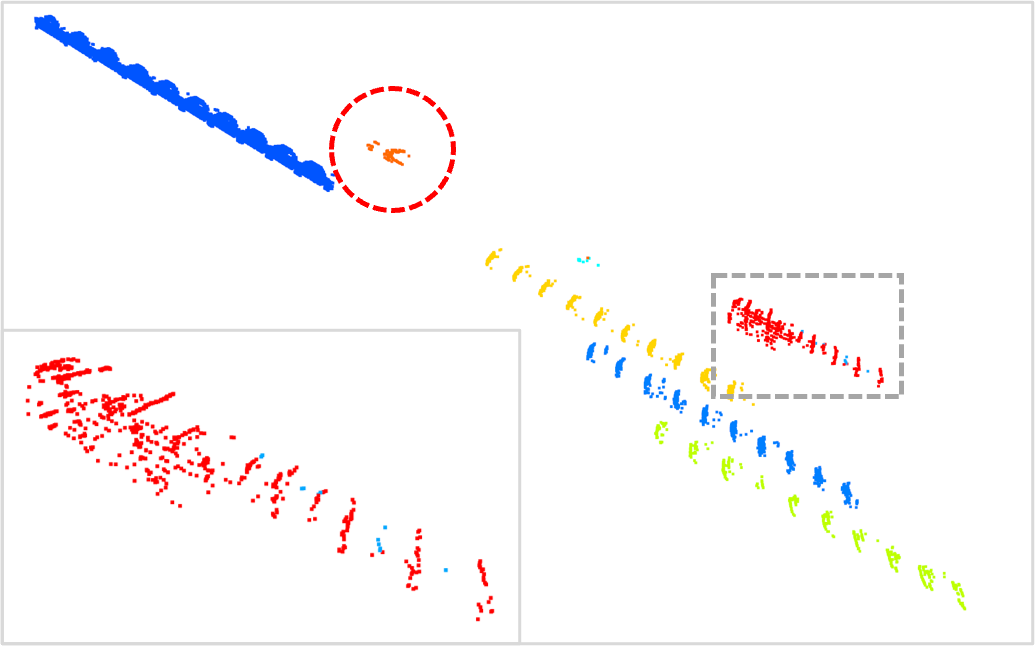}
         \captionsetup{font={footnotesize}} 
         \caption{Ours}
     \end{subfigure}
          \begin{subfigure}[b]{0.24\textwidth}
         \centering
         \includegraphics[width=\textwidth]{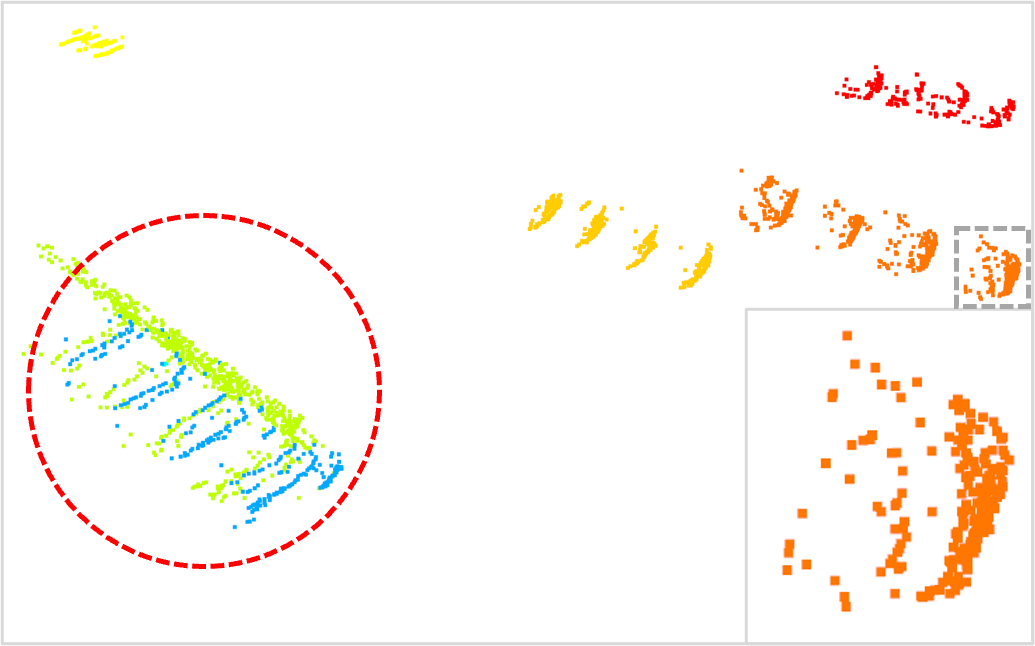}
         \hfill
         \includegraphics[width=\textwidth]{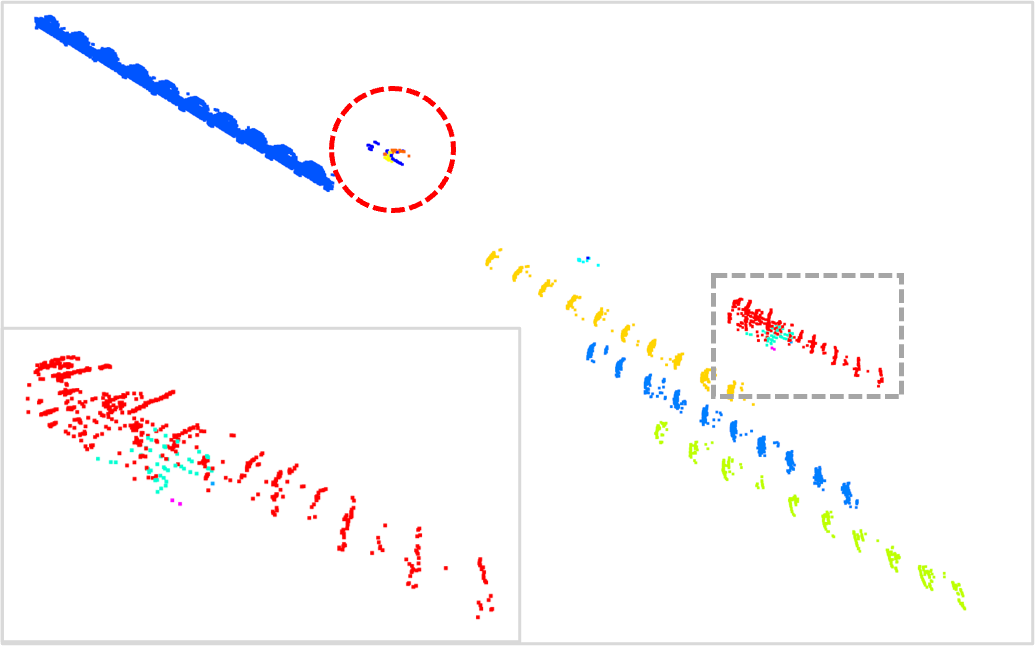}
         \captionsetup{font={footnotesize}} 
         \caption{4D-PLS \cite{aygun20214d} + PC}
     \end{subfigure}
          \begin{subfigure}[b]{0.24\textwidth}
         \centering
         \includegraphics[width=\textwidth]{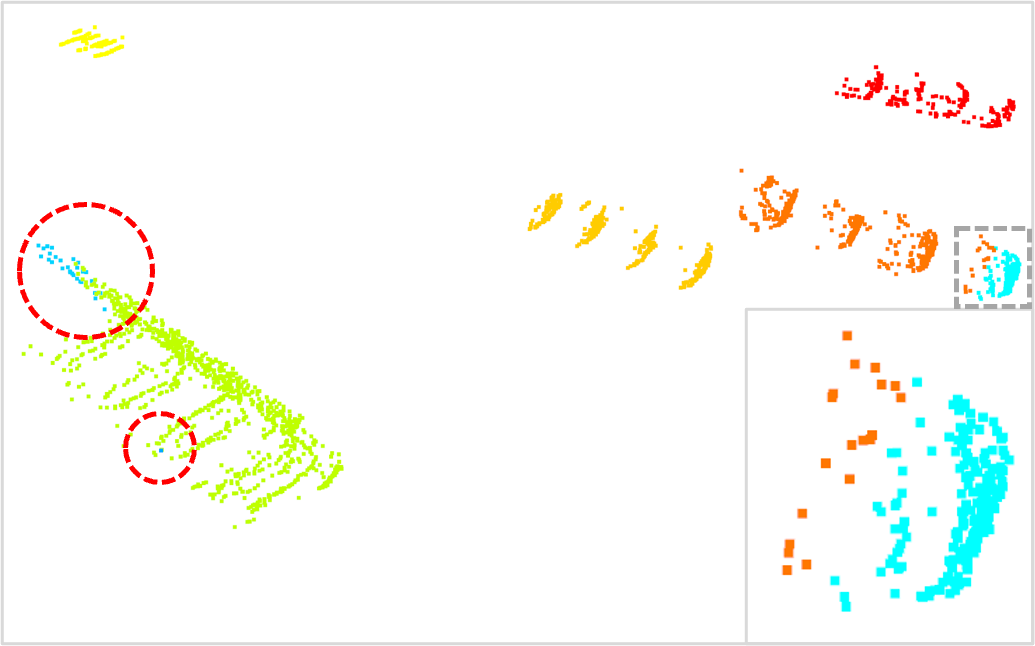}
         \hfill
         \includegraphics[width=\textwidth]{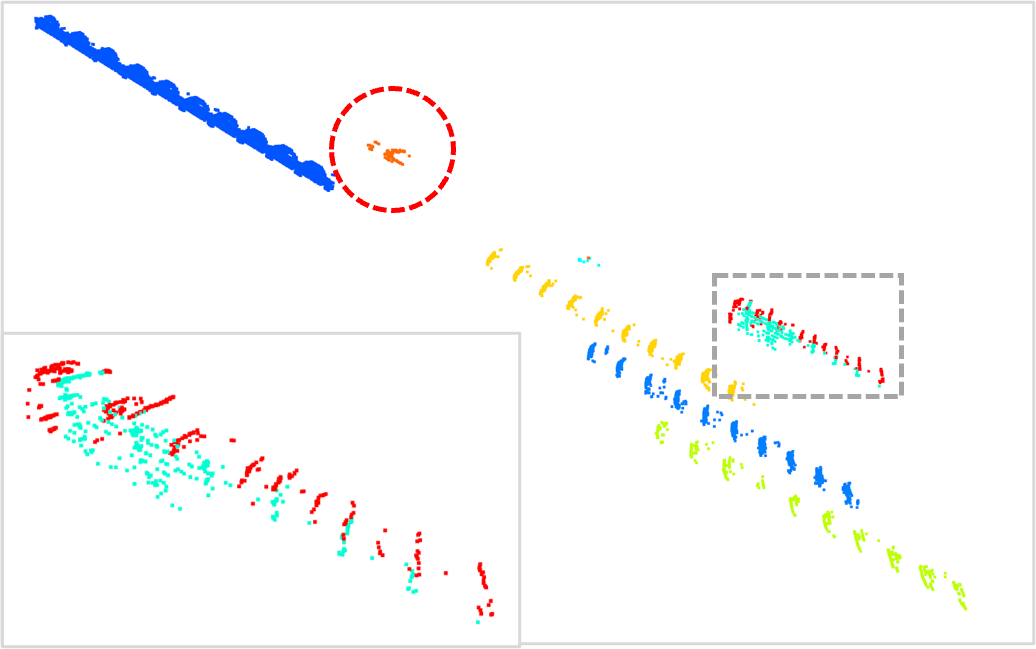}
         \captionsetup{font={footnotesize}} 
         \caption{Vanilla 4D-PLS \cite{aygun20214d}}
     \end{subfigure}
\captionsetup{font={footnotesize}} 
\caption{Qualitative comparison of our method with baseline methods. We show the tracking results for 4 frames (top row) and 10 frames (bottom row). PC means Position Compensation. The areas with significant differences are enlarged for better comparison.}
\label{results}
\end{figure*}
\subsection{Implementation Details}
\subsubsection{Ego-motion Estimation}
\begin{figure}[ht]
     \centering
     \begin{subfigure}[b]{0.18\textwidth}
         \centering
         \includegraphics[width=\textwidth]{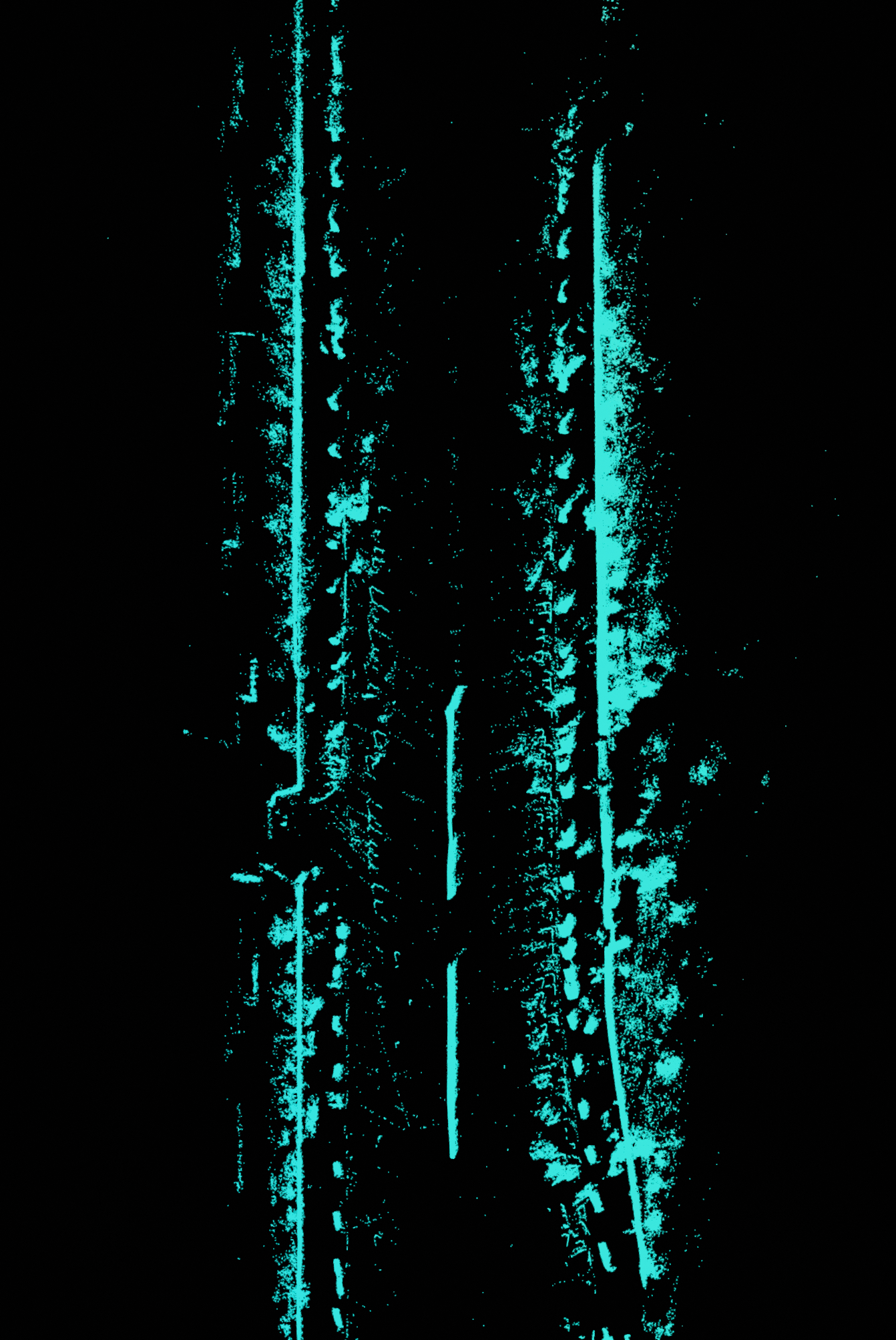}
         \caption{}
         \captionsetup{font={footnotesize}} 
         \label{SLAM}
     \end{subfigure}
     \begin{subfigure}[b]{0.29\textwidth}
         \centering
         \includegraphics[width=\textwidth]{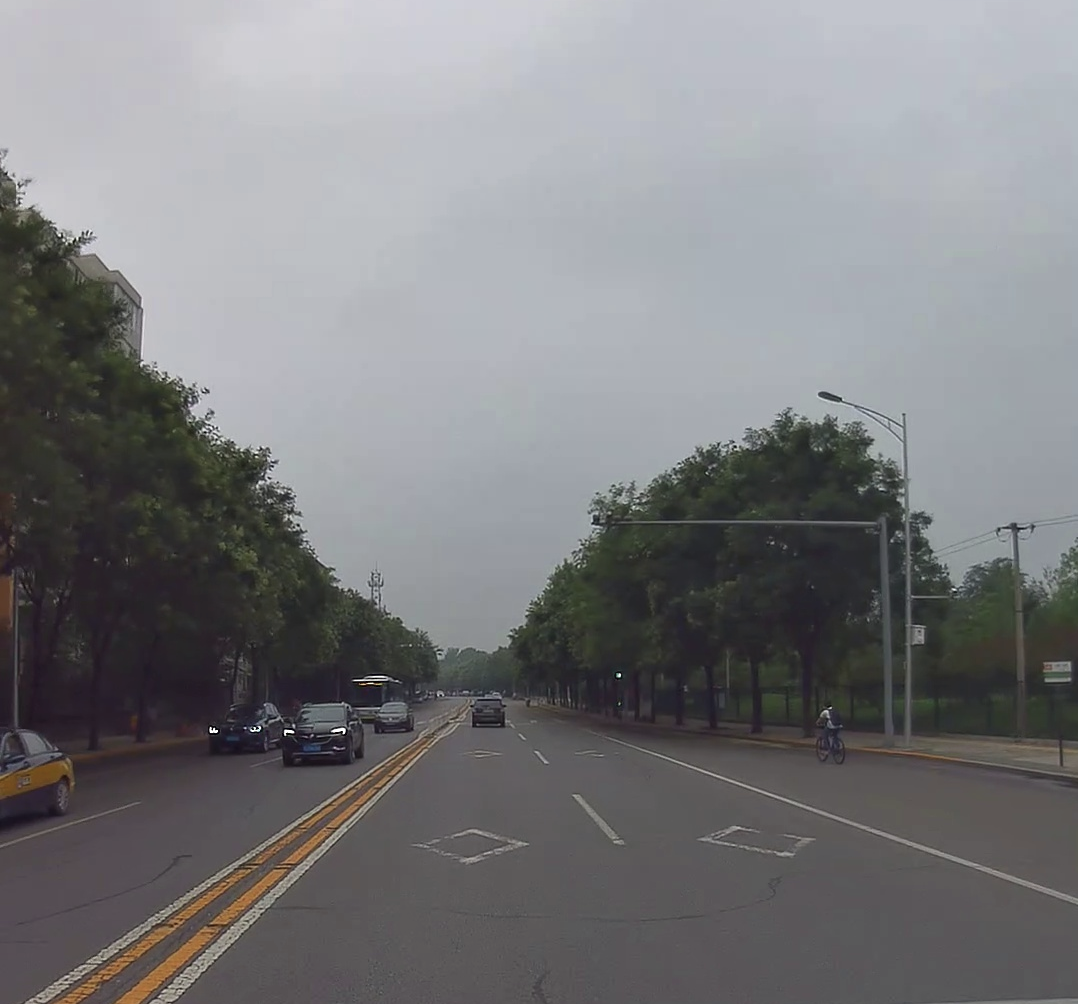}
         \caption{}
         \captionsetup{font={footnotesize}} 
         \label{image}
     \end{subfigure}
     \captionsetup{font={footnotesize}} 
        \caption{(a) is the local map by SLAM \cite{zhang2014loam} with only static non-ground points reserved. (b) is the corresponding image in front view. Note that the dynamic cars are filtered out in static map.}
        \label{map}
\end{figure}

We first filter out the outlier points as mentioned in section \ref{A}. In addition, we also remove the ground points estimated by RANSAC \cite{fischler1981random}, since the points with special pattern scanned on the ground may lead to bad registration results. We use the modified LOAM \cite{zhang2014loam} to get the ego-pose information. Specifically, we only use the static points to do the registration. We initialize the translation between the two frames by ego-velocity compensation. Other more robust registration methods, such as DICP \cite{hexsel2022dicp}, can be explored in future work. Figure \ref{SLAM} shows the local map built by our FMCW LiDAR data.

\subsubsection{Semi-Automatic Annotation}\label{annotation}
We use DBSCAN \cite{ester1996density} to cluster the moving points. The default density parameter is set as $1.2$, and the minimum number of points in each cluster is $20$. If there are less than 20 points in a cluster, we filter out these points. We double-check the results by both nearest neighbor query and cluster again after position compensation. After that, we manually check the correctness. For those failing and wrongly labeled data, we re-label them by hand.

\subsubsection{Network Detail and Loss functions}
Since the position of each cluster has been compensated, our method is not sensitive to the window size $\tau$. We validate this by choosing different window sizes (4, 6, 8, and 10) in our experiments. 

We use KPConv \cite{thomas2019kpconv} as our backbone and the complete loss function is set as:
\begin{equation}
    L = L_{SC} + \lambda_{ins}L_{ins} + \lambda_{var}L_{var} + L_{obj} + L_{reg},
\end{equation}
where $L_{reg}$ is the regression loss from KPConv \cite{thomas2019kpconv}, and $L_{var}$ is the variance smoothness loss similar to 4D-PLS \cite{aygun20214d}. We set the $\lambda_{ins}$ and $\lambda_{var}$ as $0.01$ in our experiments. And as the KPConv has provided the per-point embeddings, we can extract instance features by directly using a max-pooling operation, which is better than the other symmetric functions. 

We find that $p_{threshold}=0.4$ can get the best performance. The overlap threshold is set as $0.8$, which has a trivial effect on the final results. Note that these hyper-parameters are not changed after configuration, while the hyper-parameters in Heuristic Tracking need to be tuned scene-by-scene.

\begin{table}[htbp]
\caption{Comparison of our method with baseline methods. PC means Position Compensation.}

\label{table_performance}
\begin{center}
\begin{tabular}{c|c c c c c c c}
\hline
Method & AS & MOTSA & MOTSP & SMOTSA\\
\hline
Heuristic Tracking & 54.18 & 82.55 & 84.15 & 77.24\\
\hline
Vanilla 4D-PLS \cite{aygun20214d} & 59.55 & 80.68 & 84.87 & 76.03 \\
\hline
4D-PLS \cite{aygun20214d} + PC & 84.07 & 88.42 & 87.02 & 85.97 \\

\hline\hline
Ours & \textbf{87.89} & \textbf{89.44} & \textbf{89.07} & \textbf{89.04}\\
\hline
\end{tabular}
\end{center}
\end{table}

\subsection{Evaluation Metrics}
Since we have no class labels, we choose the following metrics \cite{voigtlaender2019mots, aygun20214d} which are only related to instance IDs, to make a fair comparison with the baseline methods: multi-object tracking and segmentation accuracy (MOTSA), multi-object tracking and segmentation precision (MOTSP), soft multi-object tracking and segmentation accuracy (SMOTSA) and association score (AS). In all of these metrics, larger values indicate better performance.

We follow the MOTS \cite{voigtlaender2019mots} to define the set of True Positives (TP), False Positives (FP), False Negatives (FN) and ID switch (IDS) to compute the MOTSA, MOTSP and SMOTSA. For AS, we follow the 4D-PLS \cite{aygun20214d} to define the TP, FP, and FN, which can be seen as a soft version of that in MOTS. In consideration that Association Score can evaluate the association in space and time in a more unified manner, we treat AS as the main evaluation indicator.

% \begin{figure}[htbp]
% {\includegraphics[width=0.11\textwidth]{res/res.png}}
% {\includegraphics[width=0.11\textwidth]{res/res1.png}}
% {\includegraphics[width=0.11\textwidth]{res/res.png}}
% {\includegraphics[width=0.11\textwidth]{res/.png}}\\
% {\includegraphics[width=0.11\textwidth]{res/res.png}}
% {\includegraphics[width=0.11\textwidth]{res/res1.png}}
% {\includegraphics[width=0.11\textwidth]{res/res.png}}
% {\includegraphics[width=0.11\textwidth]{res/res1.png}}
% \caption{Sample of our Annotations and corresponding tracking results.} 
% \label{results}   
% \end{figure}

\subsection{Results}
We compare our method with two baselines: heuristic tracking and 4D-PLS \cite{aygun20214d}. We reimplement the 4D-PLS on our dataset, without semantic output. Since 4D-PLS is sensitive to the window size, we also add position compensation to 4D-PLS to compare with our method. As for heuristic tracking, we follow our semi-automatic annotation process without human interactions. We test several sets of hyper-parameters and choose the results with the best performance. Table \ref{table_performance} shows the results of our method and baseline methods. Obviously, our method achieves the best results and generalizes well in most cases. Figure \ref{results} is the qualitative comparison of our method with baselines. Our method is more consistent in association step. It is worth mentioning that these results are not cherry-picked. 

\subsection{Ablation studies}
The velocity of points within the same instance should be close, which is also a strong cue for the association. Thus, we conduct experiments on whether to input the velocity channel. The results in Table \ref{abl_table_in} show that there is a slight improvement with velocity as input. 

Since our method has compensated for the position of each cluster, our model is not sensitive to the window size. But the window size cannot be too large, since the memory is limited and the uniform motion assumption cannot hold for large window sizes. We conduct ablation study on the window size, with $\tau = 4, 6, 8, 10$. Table \ref{abl_table_w} shows that there is no obvious difference in the performance. 
% We also validate the effectiveness of the position compensation. The results show that it is necessary to complement the position of each cluster, especially in large temporal window settings.

\begin{table}[htbp]
\caption{Ablation study on input format}
\label{abl_table_in}
\begin{center}
\begin{tabular}{c|c c c c c c c}
\hline
Method & AS & MOTSA & MOTSP & SMOTSA\\
\hline
xyz & 87.03 & \textbf{89.48} & 88.94 & 88.95\\
\hline
xyz+v & \textbf{87.89} & 89.44 & \textbf{89.07} & \textbf{89.04}\\
\hline
\end{tabular}
\end{center}
\end{table}

\begin{table}[htbp]
\caption{Ablation study on window size}
\label{abl_table_w}
\begin{center}
\begin{tabular}{c|c c c c c c c}
\hline
Method & AS & MOTSA & MOTSP & SMOTSA\\
\hline
4 scans & \textbf{87.89} & \textbf{89.44} & 89.07 & \textbf{89.04}\\
\hline
6 scans & 86.25 & 89.14 & 89.02 & 88.94\\
\hline
8 scans & 86.47 & 89.16 & 88.94 & 88.62\\
\hline
10 scans & 86.54 & 89.31 & \textbf{89.19} & 89.02 \\
\hline
\end{tabular}
\end{center}
\end{table}
% We propose a contrastive learning framework to better associate clusters from the same instance across time.

\section{CONCLUSIONS}
In this paper, we present a method of moving-object tracking based on the latest FMCW LiDAR sensor. To the best of our knowledge, it is the first work ever conducted on moving-object tracking with the FMCW LiDAR in a learning scheme. We propose a contrastive learning framework to make full use of the instance-only supervision, which can significantly improve the tracking quality. Our whole pipeline requires far less annotation effort than most existing works, e.g., MOT and MOTS, which require bounding box labels or per-point semantic labels. The results show that our method can surpass the baseline methods by a large margin. We hope our work can motivate more revolutionary works based on the FMCW LiDAR.

% \addtolength{\textheight}{-12cm}   % This command serves to balance the column lengths
                                  % on the last page of the document manually. It shortens
                                  % the textheight of the last page by a suitable amount.
                                  % This command does not take effect until the next page
                                  % so it should come on the page before the last. Make
                                  % sure that you do not shorten the textheight too much.

%%%%%%%%%%%%%%%%%%%%%%%%%%%%%%%%%%%%%%%%%%%%%%%%%%%%%%%%%%%%%%%%%%%%%%%%%%%%%%%%

%%%%%%%%%%%%%%%%%%%%%%%%%%%%%%%%%%%%%%%%%%%%%%%%%%%%%%%%%%%%%%%%%%%%%%%%%%%%%%%%

%%%%%%%%%%%%%%%%%%%%%%%%%%%%%%%%%%%%%%%%%%%%%%%%%%%%%%%%%%%%%%%%%%%%%%%%%%%%%%%%
% \section*{APPENDIX}

% Appendixes should appear before the acknowledgment.

% \section*{ACKNOWLEDGMENT}

% The preferred spelling of the word ÒacknowledgmentÓ in America is without an ÒeÓ after the ÒgÓ. Avoid the stilted expression, ÒOne of us (R. B. G.) thanks . . .Ó  Instead, try ÒR. B. G. thanksÓ. Put sponsor acknowledgments in the unnumbered footnote on the first page.

% %%%%%%%%%%%%%%%%%%%%%%%%%%%%%%%%%%%%%%%%%%%%%%%%%%%%%%%%%%%%%%%%%%%%%%%%%%%%%%%%

% References are important to the reader; therefore, each citation must be complete and correct. If at all possible, references should be commonly available publications.

\bibliographystyle{IEEEtran}
\bibliography{IEEEabrv,ref}

\end{document}